\documentclass{article}

\usepackage[nonatbib,final]{neurips_robustseq_2022}

\usepackage{subcaption}
\usepackage[justification=centering, font=footnotesize]{caption}
\usepackage[utf8]{inputenc} 
\usepackage[T1]{fontenc}    
\usepackage{hyperref}       
\usepackage{url}            
\usepackage{booktabs}       
\usepackage{amsfonts}       
\usepackage{nicefrac}       
\usepackage{microtype}      
\usepackage{xcolor}         
\usepackage{graphicx}
\usepackage{amsmath}
\usepackage{soul}

\usepackage{biblatex} 
\title{CLIFT : Analysing Natural Distribution Shift on Question Answering Models in Clinical Domain}

\author{%
  Ankit Pal \\
  Saama AI Research\\
  Chennai, India\\
  \texttt{ankit.pal@saama.com} \\
}

\begin{document}

\maketitle

\begin{abstract}
This paper introduces a new testbed CLIFT (\textbf{Cli}nical Sh\textbf{ift}) for the clinical domain Question Answering task. The testbed includes 7.5k high-quality question-answering samples to provide a diverse and reliable benchmark. We performed a comprehensive experimental study and evaluated several QA deep-learning models under the proposed testbed. Despite impressive results on the original test set, the performance degrades when applied to new test sets, which shows the distribution shift. Our findings emphasize the need for and the potential for increasing the robustness of clinical domain models under distributional shifts. The testbed offers one way to track progress in that direction. It also highlights the necessity of adopting evaluation metrics that consider robustness to natural distribution shifts. We plan to expand the corpus by adding more samples and model results. The full paper and the updated benchmark are available at \href{https://github.com/openlifescience-ai/clift} {github.com/openlifescience-ai/clift}
\end{abstract}

\section{Introduction}
Digital healthcare data has significantly expanded in recent years. Question Answering (QA) in the clinical domain involves extracting answers from clinical texts in Electronic Health Records (EHRs). Clinicians frequently consult EHR notes and documents to aid medical decision-making. \cite{Rosenbloom2011DataFC} However, developing robust clinical QA systems is challenging due to distribution shifts in real-world data. Recent studies \cite{Rawat2020, Chen2017, Devlin2019} have applied neural QA models in this domain, such as BERT and variants, with promising results. However, prior work has focused narrowly on achieving state-of-the-art accuracy on existing benchmarks rather than robust generalization. The key goal is not achieving the highest possible accuracy on a specific dataset, but developing models that can reliably transfer to new clinical data from diverse regions, hospitals, and organizations. This requires resilience to distributional shifts between populations and institutions. Therefore, progress in clinical QA should emphasize model robustness over optimization on benchmarks. Advancing the field requires rigorously evaluating model generalization under real-world distribution shifts. Our work introduces CLIFT as a benchmark designed specifically for this purpose.

In statistical learning theory, it is often assumed that test data points originate from the same distributional underpinnings as training data for a machine learning model. However, in reality, data may be sampled from various distributions with considerable domain changes \cite{Jochems2016DistributedLD, Rieke2020TheFO, Joshi2022FederatedLF}; therefore, this assumption is valid sometimes. Despite impressive results of QA models trained in the clinical domain, when data distribution unexpectedly changes, models that seem to be doing well in test set accuracy fail when deployed in real-time. \cite{Yue2020ClinicalRC} research indicates that when predicting questions that have been paraphrased and completely new, the F1 score drops by around 40\% to 60\%. We are evaluating the Models trained on emrQA to investigate natural distribution shifts for different test sets, for example, changing from an existing test set to a cancer, heart, and other test sets.
Our Study focuses on a question:

\textit{Are Clinical Models fine-tuned on emrQA robust to natural distribution shifts?}

We perform a comprehensive experimental study on the emrQA dataset \cite{Pampari2018} in this paper. We introduce a new testbed derived from the MIMIC-III database. Our results show that emrQA models exhibit robustness to some natural distribution shifts. Despite impressive results on the training dataset, when applied to new test sets, the performance degrades, and models in our testbed experiences a drop in F1 metric. Our experiments underline the need and the potential for improving the robustness of clinical domain models under distributional shifts. The results also highlight the necessity of adopting evaluation metrics that consider robustness to natural distribution shifts.

In brief, the contributions of this study are as follows. (i) We are proposing five new test datasets. The testbed covers different clinical domain diseases and multiple sub-topics. These questions are from MIMIC-III \cite{Johnson2016}, and human experts validated all the questions, answers, and evidence triples. (ii) This dataset offers several advantages over existing datasets. They evaluate the diverse reasoning abilities of clinical models over a vast array of disease topics and subtopics. The testbed is a comprehensive evaluation of clinical domain data.
(iii) Detailed statistics and fine-grained evaluation of natural distribution shift are provided in this study. We examine and quantify the shortcomings of the clinical deep learning models trained on emrQA when evaluated with new test data. Moreover, We also analyzed the dataset on difficulty metrics such as answer diversity and Syntactic divergence. (iv) We conduct our experiments with the help of the HuggingFace Transformers library \cite{Wolf2020}. Furthermore, we also open-source the code and model checkpoints to reproduce the results and test with new models in the future. (v) We provide a new QA tool to validate the Question Answering Dataset. We hope this will accelerate the tagging process and robust ML research in this domain.

\begin{figure*}[!h]
    \centering
  \includegraphics[width=14cm]{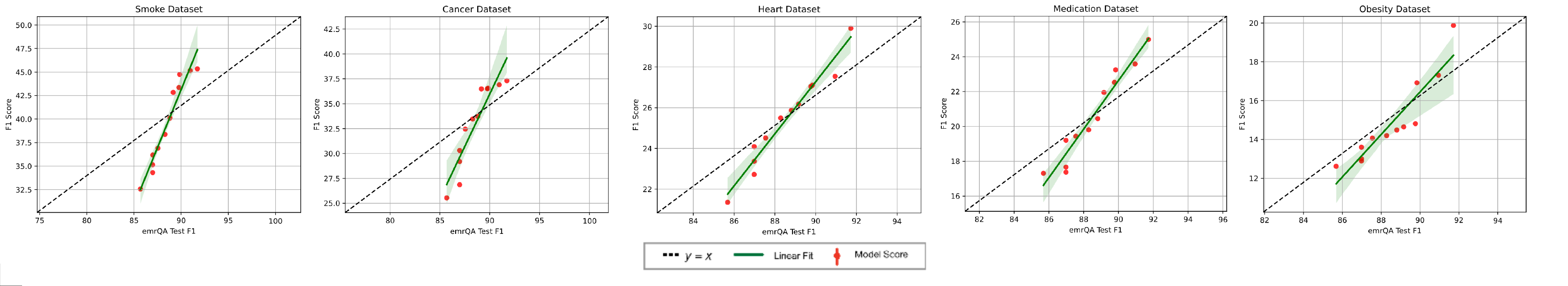}
  \caption{ \footnotesize Comparison of models’ F1 scores on emrQA and our proposed test datasets. The result demonstrates that despite impressive results on the training dataset when applied to new test sets, the model’s performance degrades, which shows the distribution shift}
  \label{fig:shiftsanalysis}
\end{figure*}
\section{Background}
\subsection{Natural Distribution shift}
For a prediction algorithm $\mathcal{M}$ : $\mathcal{X}$ $\rightarrow$ $\mathcal{Y}$ We assume a clinical domain dataset from a Hospital $A$ where input space is $X$ and label space is $Y$. To evaluate $M$'s performance on a test dataset $Q=\{x_i|i=1,\dots, n\}$ drawn from a data distribution $P$, we will assume that algorithm $M$ is fixed. The initial steps would be choosing a loss function $\ell: \mathcal{Y}\times\mathcal{Y} \rightarrow \mathbb{R}$ and calculating the loss $\ell_Q(M)=\sum_{i=1}^n \ell\left(M\left(x_i\right), y_i\right)$ and population loss $\ell_{\mathcal{P}}(M)=\mathbb{E} \mathcal{P}[\ell(M(x), y)]$ under the test distribution.While the input $X$ and label $Y$ spaces are fixed in this case, there may be many distinct joint distributions. Furthermore, Besides the dataset from Hospital $A$, we collect another clinical domain dataset from Hospital $B$ where the test set is $Q^{\prime}$ from a new distribution $P^{\prime}$

In addition to the test loss $l_Q(M)$, we are interested to know the loss $l_{Q^{\prime}}(M)$ under real-time applications, known as distribution shift. The observations of models' expected loss to changes or differences would be helpful to improve the model's robustness, eliminate the biases and understand the dataset better. This difference may be broken down into three parts.\cite{Miller2020}
\begin{equation}
\small
l_Q(M)-l_{Q^{\prime}}(M)=\underbrace{\left(l_Q(M)-l_{\mathcal{P}}(M)\right)}_{\text {Adaptivity gap }}+\underbrace{\left(l_{\mathcal{P}}(M)-l_{\mathcal{P}^{\prime}}(M)\right)}_{\text {Distribution gap }}+\underbrace{\left(l_{\mathcal{P}^{\prime}}(M)-l_{Q^{\prime}}(M)\right)}_{\text {Generalization gap }}
\end{equation}


The adaptivity gap quantifies how adjusting the model to the test set affects the estimate of population loss.
While the distribution gap indicates how much a change in the sample's distribution impacts the model's performance, the difference between the sample and population losses due to $Q$'s random sampling is captured by the generalization gap.

\section{emrQA Dataset}

emrQA \cite{Pampari2018} is a large-scale extractive question-answering dataset generated semi-automatically from question templates, and human experts annotated the templates. The proposed task was to extract a continuous text span from the clinical note while giving a clinical note and a question as input to the QA model. The corpus contains almost 400,000 triplets of a question, logical form, and evidence/answer. 

We used Medication and Relation datasets for fine-tuning the models. The following two comprise 80\% of the overall emrQA dataset, and their structure is congruent with the span prediction task, which is more significant for clinical decision assistance. Table \ref{tab:emrqadata} describes the statistics of the emrQA dataset.

\section{Proposed Testbed}

The MIMIC-III \cite{Johnson2016} is the most popular dataset for Electronic Health Record (EHR) related tasks. In the dataset, A clinical record is maintained for each patient, which contains information such as ECGs, physician notes, discharge summaries, nurse progress reports and more. Due to its comprehensive collection of unstructured clinical notes, MIMIC-III has been extensively used in clinical NLP tasks.\cite{Park2021KnowledgeGQ}

In order to evaluate the ability of healthcare domain QA systems to generalize to new clinical data, we created five new test sets from MIMIC-III. First, we randomly sampled the patient's records from the MIMIC-III dataset, and later, we filtered the sampled data based on the Heart, Medication, Obesity, Smoking, and Cancer subset's keywords. Finally, we utilized the same question templates provided by emrQA to create the question and answer pairs, along with a few more templates annotated by experts.

\textbf{Heart Disease Test set} The Heart disease test subset contains 1.5k annotated questions and answers related to heart disease and issues such as heart attack, heart failure, coronary artery disease Ischemia, and more.

\textbf{Medication Test set} The medication subset contains around 1.5k samples. It includes the question and answers related to medical prescriptions, drugs name, drug dosages, history, and \& status of medications; it also contains questions related to the number of doses, side effects of medications, reasons for medications and more.

\textbf{Obesity Test set} Similar to the above datasets, we extracted 1k questions and answers related to obesity, such as comorbidities associated with the current issue, hypertension, Sleep Apnea, Diabetes, and more. The test set's diversity helps to quantify the degree to which current systems overfit the original test set.

\textbf{Smoking Test set} For the smoking subset, 2k questions were selected related to the current status \& history of smoking, quantity \& type of smoking, issues and more.

\textbf{Cancer Test set}
The subset includes 1.5k questions and answers related to different cancers, such as Breast Cancer, Colon cancer, Rectal Cancer, Thyroid Cancer, and more. Moreover, the set also contains samples mentioning the Cancer cause, side effects, medications for current cancer, stage of cancer and more.

\subsection{Preprocessing \& Quality Checks}

The following procedures were used to sanitize the raw data and to ensure that all questions could be answered using textual input. We lowercase the text in the subset test sets, remove punctuation marks and tokenize the text using whitespace. We set the min length to 100 and split the long documents into sub-documents. We also converted a few abbreviations to complete forms such as y.o., y/o, to a year old and yr, yr's, yrs to years. Also, additional data cleaning procedures were performed to ensure that the information gathered from the question, answer, and evidence triples met the data quality objectives. We generated 25k question-answers pairs from selected medical records; 15k samples were discarded due to low data quality, noisy and wrong answers. After the data team re-validated, at the time of writing the paper, 7.5k final samples were finalized for the testbed. However, We plan to expand the corpus by adding more samples and model results. The full paper and the updated benchmark are available at \href{https://github.com/openlifescience-ai/clift} {github.com/openlifescience-ai/clift}

\subsection{Dataset Statistics \& Analysis}

This dataset covers different diseases and medical conditions based on patient information.

Table \ref{tab:datasetstats} in the Appendix summarises the general statistics of preprocessed data. Moreover, Figure \ref{fig:tokendist} in the Appendix depicts an additional useful statistic: the number of unique tokens in the dataset. The performance of the models is influenced by the amount of vocabulary, which is a good indicator of the linguistic and domain complexity associated with a text corpus. The Obesity \& Heart subset was observed to include more extended questions and a broader vocabulary than the others. As a result, the former subsets' question are more complex than the rest.

\begin{figure*}[!h]
\centering
  \includegraphics[width=8cm]{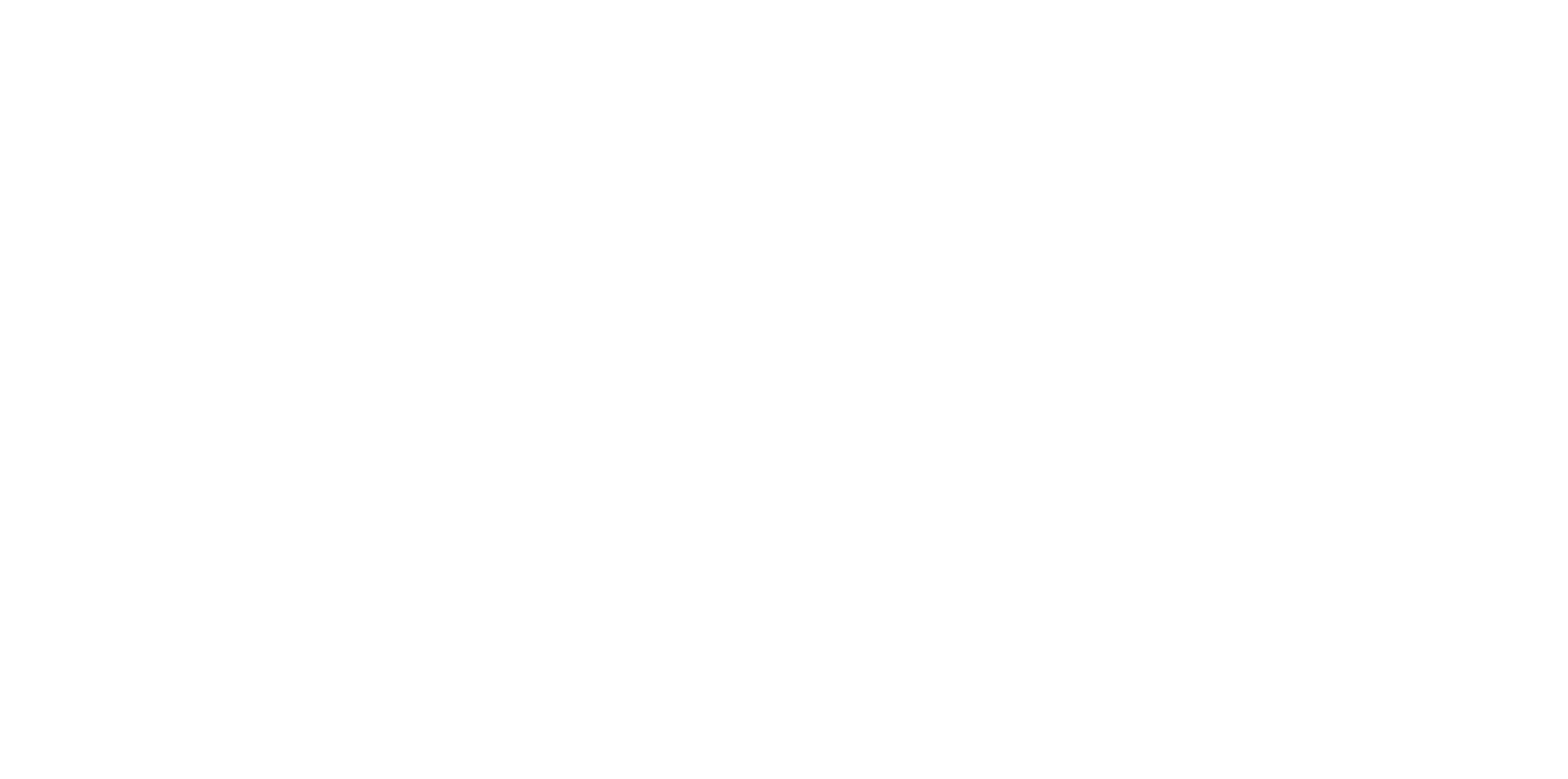}
  \caption{ \footnotesize Syntactic divergence histograms for the emrQA \& proposed datasets' question and response phrases.
}
  \label{fig:analysisstats}
\end{figure*}

In addition, We use the two difficulty metrics provided by \cite{rajpurkar-etal-2016-squad} to compare the original emrQA test set to our five new test sets. Figure \ref{fig:analysisstats} shows the Syntactic divergence histograms for the emrQA \& proposed datasets.

\section{Experimental Setup}

We followed the \cite{Yue2020ClinicalRC} random seed and split the combined Medication and Relation datasets based on the clinical texts obtaining train, dev, and test with the ratio 7:1:2. Table 7 shows the descriptive statistics of the emrQA dataset. Similar to \cite{rajpurkar-etal-2016-squad}, we used the F1 score (F1) as our evaluation matrices. The hyperparameters are chosen based on \cite{Devlin2019}. In order to fine-tune it for four epochs, we set the batch size to 8, the learning rate to 2e5, and the maximum response length to 30. we optimize the model's parameters using the AdamW optimizer.

We implemented these models using Pytorch \cite{Paszke2019PyTorchAI} and obtained pre-trained checkpoints from the HuggingFace library \cite{Wolf2019HuggingFacesTS}. All models are fine-tuned on 4 GPUs with NVIDIA Quadro RTX 8000 and 256 GB of RAM.

\subsection{Baseline Models}
\label{baselinemodels1}

In our baseline studies, we consider 12 existing models as a baseline, which have shown strong performance in biomedical NLP tasks. These are developed on several pre-trained language models employing the Transformers architecture \cite{NIPS2017_3f5ee243}. Notably, BioBERT \cite{Lee2019}, and variant BioClinicalBert and BioBert-squad-L pre-trained on large-scale biomedical corpora. DistilBert is a distilled version of the BERT base model. We used two different versions of DistilBert, including DistilBert-squad-B and BERT-ClinicalQA, fine-tuned on Squad and ClinicalQA, respectively.

To better utilize the knowledge of MIMIC and PubMed documents for QA tasks, we fine-tuned the BlueBert Models, including BlueBert-MIMIC and BlueBert-PubMed. Moreover, We also used the Roberta architecture models, including Biomed-Roberta, and Roberta-squad-B. In addition, we fine-tuned the PubMedBERT, Bert-Large, and Electra-Squad-L on the emrQA dataset. Section \ref{appendix:models} in the Appendix describes the models in detail.

We finetuned the models on the training dataset of emrQA and evaluated on the proposed test sets.

\section{Results \& Discussion}

The objective of the proposed testbed is to test the performance of a clinical model when evaluated on data relevant to the same Question-Answering task but from multiple data sources/hospitals and environments.
We ran a battery of experiments and evaluated twelve models on proposed datasets. In addition to investing the state of art models in clinical domains for Question Answering, we examine the robustness of models on the natural distribution shift in the emrQA dataset. More information about baseline models is provided in Section \ref{baselinemodels1} and Appendix \ref{appendix:models}

\subsection{Comparison of Transformer Based clinical Models on emrQA \& Clift}

Under the emrQA test set, PubMedBERT performed better than other models. Moreover, Biomed-Roberta gave the second-best results on the same. The Roberta-squad-Base model outperformed other models on the Medication, Obesity, and Heart test set. However, Bert-Large BlueBert-MIMIC gave the best accuracy on Cancer and Smoke datasets.

To examine the QA model's result closely, We also performed interpretability experiments and provided the results in Appendix \ref{appendix:interpretabilityresult}

\subsection{Are Clinical Models fine-tuned on Emrqa robust to natural distribution shifts?}

To test the hypothesis that the original test set performance is better than new test sets, we compare the F1 scores for all the models trained on the emrQA to the F1 scores on each of our additional test sets. Figure \ref{fig:shiftsanalysis} shows that, when applied to proposed test datasets, All the models indicate a performance loss ranging from 43.82\% to 73.07\%. Table \ref{tab:medicationtab} in the Appendix shows the performance of different models on the medication test set. As we can see that BERT-ClinicalQA drops around 70\% F1 points while evaluating the medication dataset, compared to a drop of about 61\%, 55\%, 75\%, and 53 \% F1 points when evaluating the model on heart, cancer, obesity, and smoke dataset respectively.

PubMedBERT fine-tuned on emrQA shows the worst performance on the Heart test set with a gap of 70.37 \% F1. Moreover, Biomed-Roberta shows a large gap of 64.22 \% on the Cancer dataset
and DistilBert-squad-B, BioBert-B suffers from generalizing well on the Obesity and Smoke dataset and dropping by 73.07 and 54.98 respectively, in the performance. Roberta-squad-B outperformed all other models on 3 out of 5 datasets, including Medication, Heart, and Obesity, with an F1 score of 24.99\%, 29.90\%, and 19.88\%, respectively.

These results suggest that the models are sensitive to distribution shifts. Real-world deployments of health machine learning systems may encounter these failure situations. Therefore, depending on the entity of interest, performance for downstream applications would vary greatly. Table \ref{tab:cancertab}, \ref{tab:obesitytab}, \ref{tab:hearttab} and \ref{tab:smoketab} in the Appendix show the results of different models on cancer, obesity, heart, and smoke dataset respectively.

These results imply the potential for further study handling distribution shifts in clinical datasets. In addition, future research should look at specific model training parameters and the datasets' size, quality, and quantity that contribute to these variations.

\section{Conclusion}

Robust machine learning aims to create techniques that consistently work in different environments. To this end, we propose a CLIFT testbed to evaluate the clinical deep learning model's performance under the distributional shift of the different test sets. We evaluate this testbed on several algorithms trained on the emrQA dataset and find that the performance degrades when applied to new test sets despite impressive results on the original test set. Moreover, In this study, we performed an empirical analysis on emrQA and analyzed the proposed test sets on difficulty metrics such as answer diversity and syntactic divergence. Our extensive testbed will indicate progress towards trustworthy machine learning in healthcare.

\section*{Acknowledgments}
We thank Samsudeen Mugaideen for building essential Front-end tools for data tagging and validation. We greatly appreciate the support of our lovely team members with medical backgrounds Srishti Awasthi, Amanpreet Kaur, Dipali Mishra, Chitvan Gautam and Nidhi, who played a crucial role in validating the tagged data, supported by Geetha Eswaran and Prashant Shukla. We also want to thank anonymous reviewers \& Malaikannan Sankarsubbu for their valuable comments \& feedback.  

\printbibliography 







\appendix

\section{Evaluation Metric}

We use the following metric to measure the performance of the clinical question-answering models. Assuming the clinical text passage is ${t}$, a question is denoted by ${q}$, answers are denoted by ${a}$, and the dataset sample and the model are represented by ${X}$ and ${f}$ respectively.

\subsubsection{F1-Score}
F1 score is based on individual word level. It is a standard metric for classification tasks and is widely utilised in Question Answering. The F1 score measures the average overlapping between a model's output and ground-truth span. It is a weighted average of precision and recall. The ratio of shared words to the total number of words in the model's output is called precision, while the ratio of shared words to the total number of words in the ground truth is called recall.

\begin{equation}
\mathrm{F} 1 Score(f)=\frac{1}{|X|} \sum_{\left(t, q,\left(a^1, \ldots, a^n\right)\right) \in X} \max _{i=1, \ldots, n} k\left(a^i, f(t, q)\right)
\end{equation}

Where $k$ is the harmonic mean of precision and recall between the
two sets.

\section{Interpretability Analysis}
\label{appendix:interpretabilityresult}
In order to understand the attention layer and prediction on the dataset, we investigated the output from a model trained on the emrqa dataset. In addition, we utilized the interpretation hooks to examine and better understand embeddings, sub-embeddings and attention layers.

\subsection{Probability Mass Function}
\begin{figure*}
\centering
  \includegraphics[width=12cm]{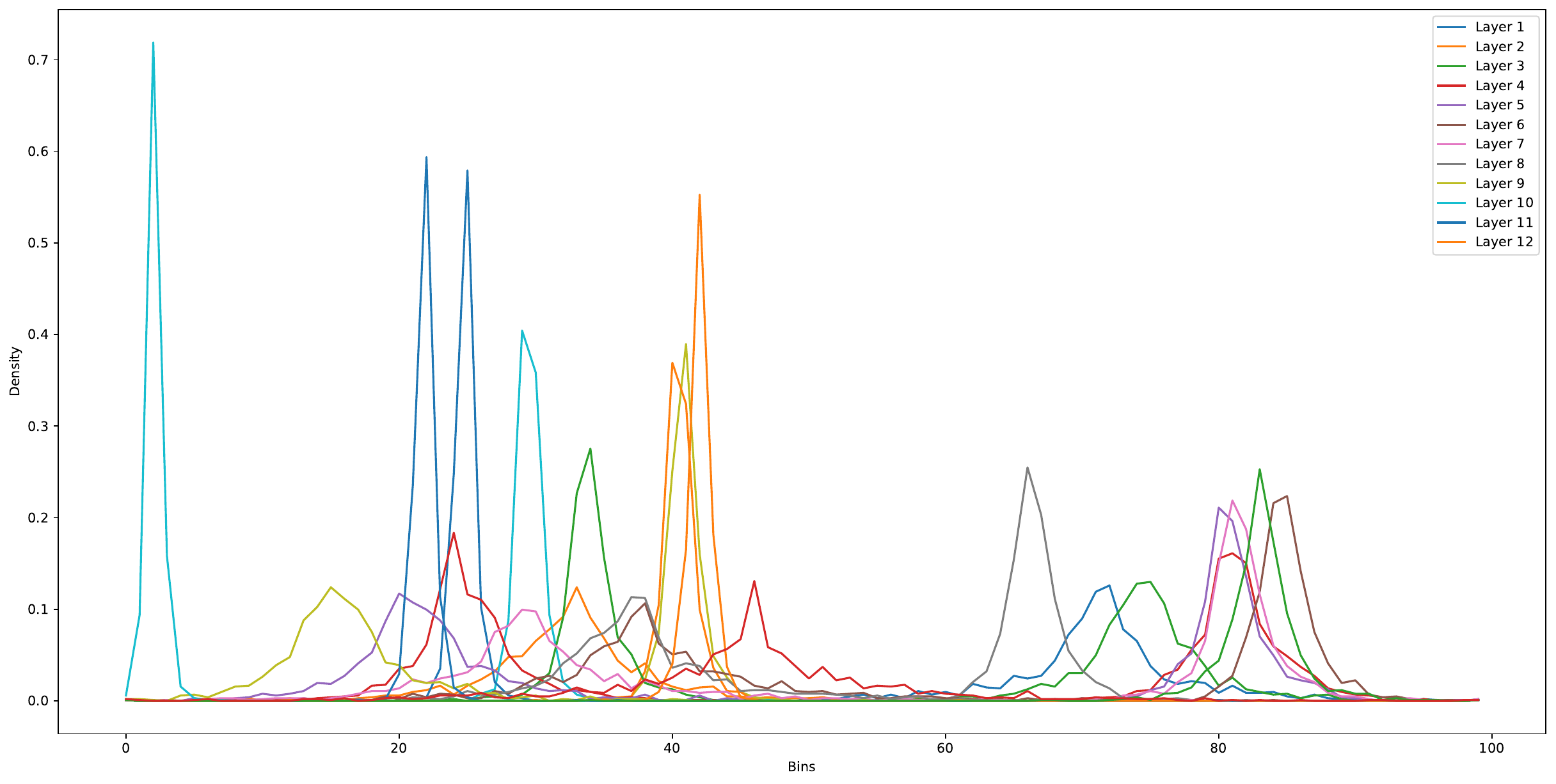}
  \caption{ \footnotesize Visualization of the probability mass function of attributions for end position tokens.
}
  \label{fig:pmfgraph}
\end{figure*}

To understand and compare the distributional attributions patterns across multiple layers, we visualized each layer and presented them in Figure \ref{fig:pmfgraph}.

In order to calculate the entropy of each layer's attributions, we represented the attributions as a probability density function (pdf) and estimated the entropy of that. The plot illustrates the Probability Mass Function (PMF) of each layer's attributions for the start and end position tokens. We can see from the figure that the distributions are taking on bell-shaped forms with various means and variances.

\subsection{Attributions HeatMap}
We used the layer conductance algorithm to visualize and understand the distribution of attribution scores for tokens across all layers.

\begin{figure*}
\centering
  \includegraphics[width=12cm]{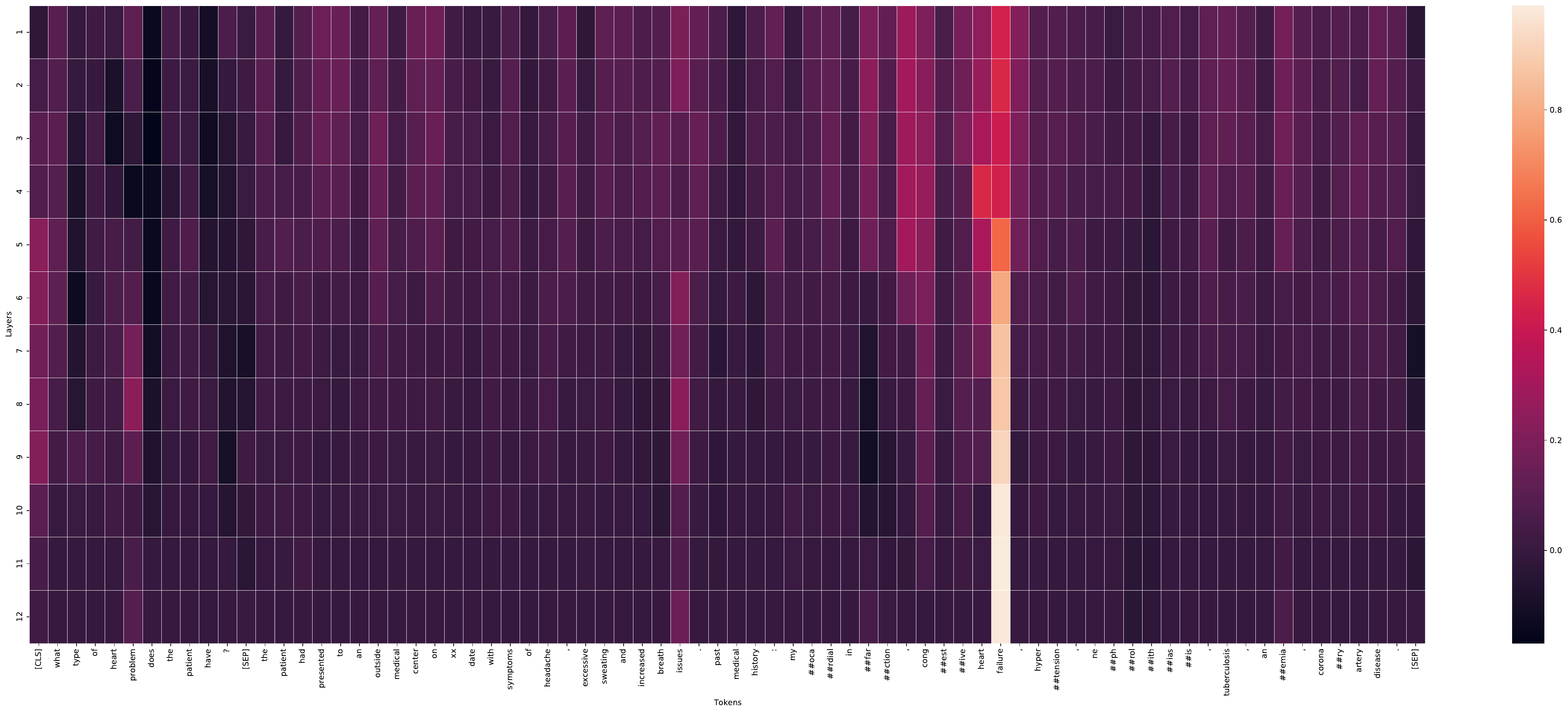}
  \caption{ \footnotesize Heat map of all layers' tokens \& attributions for the start position prediction. 
}
  \label{fig:heatmapplot}
\end{figure*}

Figure \ref{fig:heatmapplot} shows the heat map of end position tokens across all layers. Interestingly, the word heart failure gains high attribution from layers seven to twelve and one to twelve respectively. Conversely, many other tokens have negative or close to zero attribution in the first six layers.

\subsection{Distribution of attributions per layer}
\begin{figure*}
\centering
  \includegraphics[width=12cm]{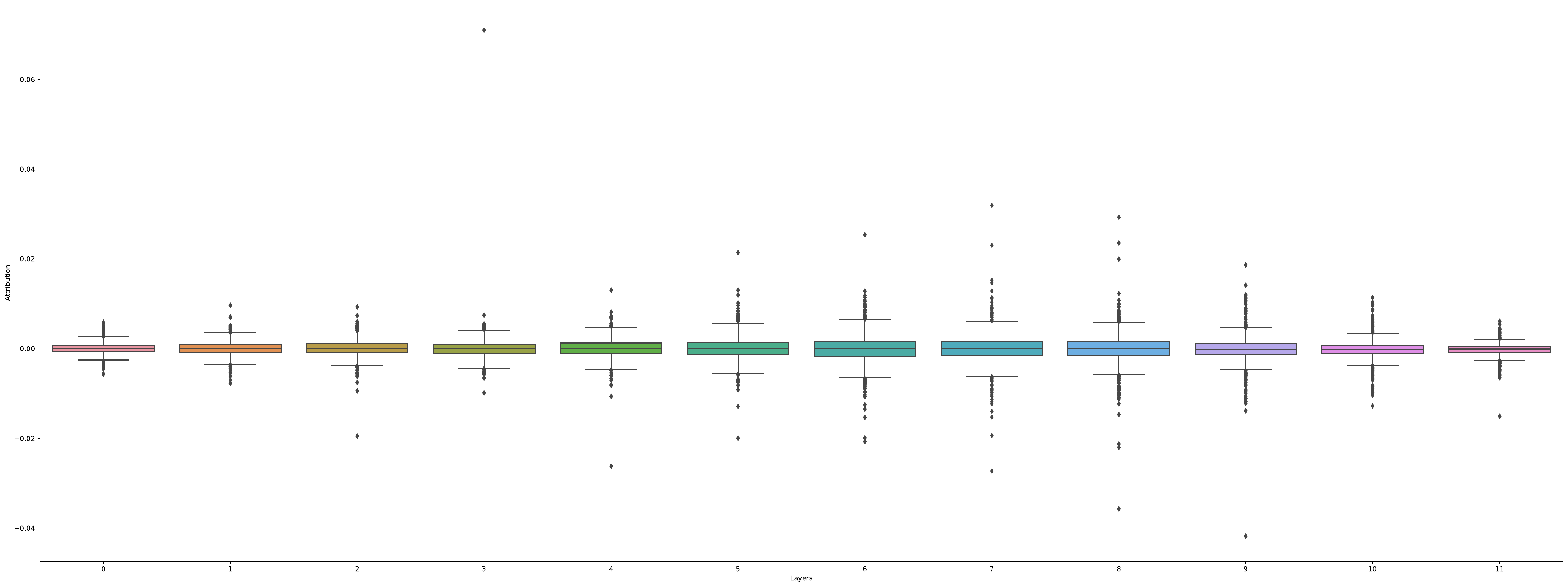}
  \caption{ \footnotesize Distribution of attributions per layer in start position.
}
  \label{fig:outstart}
\end{figure*}

Figure \ref{fig:outstart} shows the end position's tokens box plot. From the diagram, one can see the presence of outliers from 4 to 9. We also observe that the interquartile range slowly decreases for start position prediction as we move deeper into the layers and finally diminish.

\begin{figure*}
\centering
  \includegraphics[width=12cm]{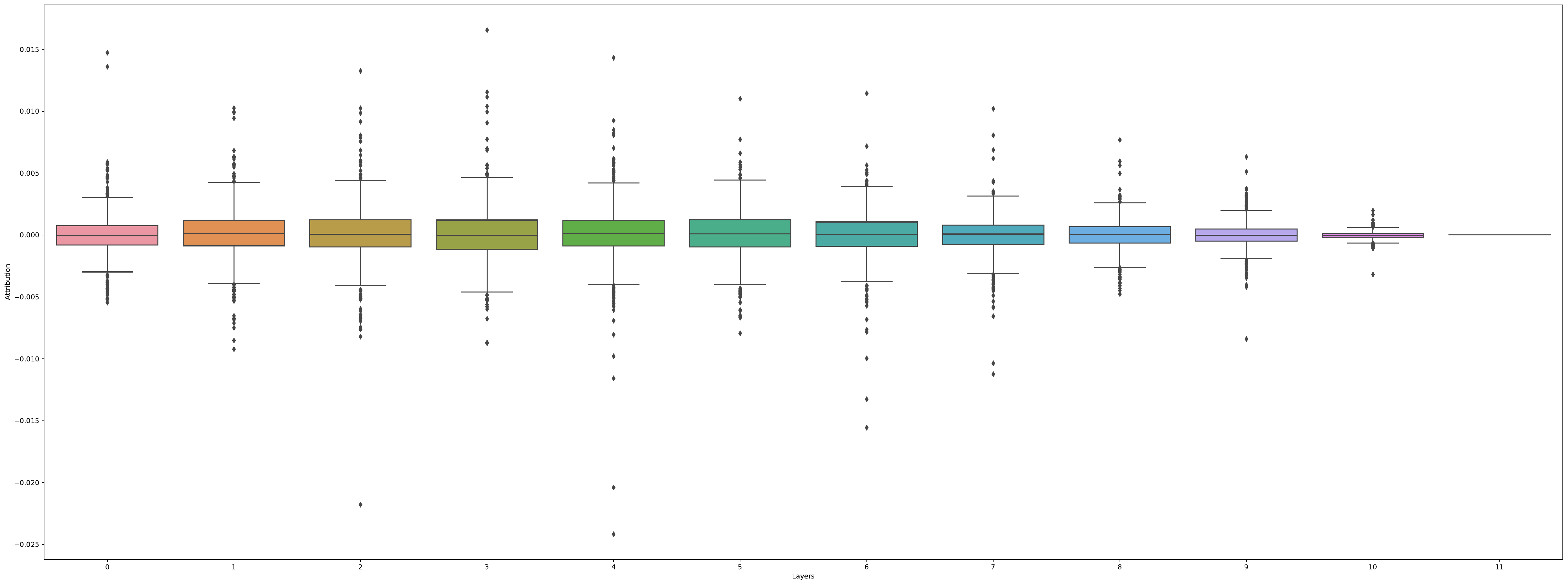}
  \caption{ \footnotesize Distribution of attributions per layer in end position.
}
  \label{fig:outend}
\end{figure*}

We also visualised the prediction of the start position's distribution box plot. Figure \ref{fig:outend} shows the outliers from layer 1 to layer 7. The interquartile range stays mostly the same when moving deeper into the layers.

\section{In-depth Analysis}
\label{appendix:graph}
Based on the outcomes of our experiments and model evaluation, we can say that the proposed testbed is challenging \& complex.
In order to support this argument \& take a closer look at the proposed test sets, We use the difficulty metric provided by \cite{rajpurkar-etal-2016-squad} to compare the original emrQA test set to our four new test sets.

\subsection{Syntactic divergence}

Syntactic divergence assesses how similar the syntactic dependency tree structures of the question and response phrases are. Figure \ref{fig:analysisstats} shows the syntactic divergence histograms for our new test sets and the emrQA test set.

\section{Model Details}
\label{appendix:models}
\textbf{BioClinicalBert}
The MIMIC III database, which contains EHR from ICU patients at the Beth Israel Hospital in Boston, Massachusetts, was used to train the Bio ClinicalBERT model \cite{Alsentzer2019PubliclyAC}. The NOTEEVENTS table's whole collection of notes (880M words) was included.

\textbf{BioBert-B} stands for Biomedical Bert, a domain-specific language representation model pre-trained on large-scale biomedical corpora. \cite{Lee2020BioBERTAP} shows that in many biomedical text mining tasks, BioBERT performs much better than BERT and earlier state-of-the-art models.

\textbf{BioBert-squad-L} The large variant of the BioBert model, was trained on the same PubMed abstracts with more parameters \& domain-specific vocabulary based on WordPiece \cite{Lee2020BioBERTAP}. We utilised the Large BioBert model, which has been finetuned on the Squad dataset for the question-answering task.

\textbf{BERT-ClinicalQA} DistilBERT \cite{Sanh2019DistilBERTAD} is the distilled version of the BERT base model , and it is smaller and faster than BERT, which was pre-trained on the same corpus using the self-supervised mechanism. It uses the BERT base model as a teacher. We used the DistilBERT version, which has already been fined-tune on the clinical QA dataset.

\textbf{DistilBert-squad-B} is the base variant of DistilBERT \cite{Sanh2019DistilBERTAD} and it was fine-tuned using knowledge distillation on the SQuAD dataset.

\textbf{BlueBert-MIMIC} is a Transformer based Bert model \cite{Peng2019TransferLI} pre-trained on clinical notes from the MIMIC-III dataset. The result presented in the paper shows that Bluebert outperforms Bert and other general domain models on the BLUE benchmark, consisting of five tasks and ten datasets covering both biomedical and clinical literature with varying dataset sizes and difficulty.

\textbf{BlueBert-PubMed} Similar to the BlueBert-MIMIC model, BlueBert-PubMed \cite{Peng2019TransferLI} was trained on PubMed abstracts. The final training dataset contains around 4,000 million Words.

\textbf{Biomed-Roberta} is a variant of RoBERTa-base \cite{Liu2019RoBERTaAR} architecture. The authors utilised the continued pretraining approach to train the model on 2.68 million scientific papers from the Semantic Scholar. \cite{Gururangan2020DontSP} The training dataset consists of the full text of the papers with 7.55B tokens and 47GB of data in total.

\textbf{Roberta-squad-B} This is a Roberta Base architecture fine-tuned on the SQuAD2.0 dataset \cite{Rajpurkar2018KnowWY}. It has been trained for Question Answering Task on question-answer pairs dataset, including unanswerable questions.

\textbf{PubMedBERT} is based on Bert's architecture \cite{Gu2022DomainSpecificLM}.The Model has been pre-trained on the full articles from PubMedCentral and abstracts from Pubmed from scratch. The results show that it outperforms different Transformer based models on Biomedical NLP tasks.

\textbf{Bert-Large} The Model is based on transformer-based architecture. First, it was trained with a technique called Word masking. The word masking technique masks all the tokens that correspond to a word while the overall masking rate remains constant. After the pre-training process, the Model was fine-tuned with the SQuAD dataset.

\textbf{Electra-Squad-L} The model \cite{Alrowili2021BioMTransformersBL} was pre-trained on abstract data from PubMed, fine-tuned on the SQuAD2.0 dataset, and later fine-tuned on the BioASQ8B-Factoid training dataset. The result shows that fine-tuning helped the model to achieve a better score on BioASQ challenge Benchmark.

\begin{table*}
\small
\centering
\resizebox{0.40\textwidth}{!}{%
\begin{tabular}{lccc}
\toprule
 & {\bf Medication} & {\bf Relation} & {\bf Total}\\
\midrule

Question \# & 222,957 & 904,592 & 1,127,549\\ 
Context \# & 261 & 423  & 684\\
Template \# & 80 & 139 & 219\\
Avg Q tokens & 8.00 & 7.91 & 7.9\\
Avg A tokens & 9.47 & 10.41 & 9.9\\
Avg C tokens & 1062.66 & 889.23 & 975.9\\

\bottomrule
\end{tabular}}
\caption{emrQA datasets' statistics, where Q, A, and C represent the Questions, Answers and Context respectively.}
\label{tab:emrqadata}
\end{table*}

\begin{table*}
\small
\centering
\resizebox{0.60\textwidth}{!}{%
\begin{tabular}{lccccc}
\toprule
 & {\bf Smoke} & {\bf Heart} & {\bf Medication} & {\bf Obesity} & {\bf Cancer}\\
\midrule
Question \#  &  2k & 1.5k  & 1.5k  & 1k & 1.5k\\
Min Q tokens & 2 & 3 & 3 & 3  & 3\\
Min A tokens & 1 & 2 & 1& 1 & 1\\
Min C tokens & 100 & 200& 100& 100 & 100\\
Avg Q tokens & 6.42 & 8.31 & 7.61& 7.19 & 8.40\\
Avg A tokens & 4.59 & 4.15 &      3.93 &   4.04 &  4.12\\
Avg C tokens & 217.33 & 234.18 & 215.49& 212.88 & 210.16\\
\bottomrule
\end{tabular}}
\caption{Proposed datasets' statistics, where Q, A, and C represent the Question, Answer, and Context, respectively.}
\label{tab:datasetstats}
\end{table*}

\begin{table*}
\small
\centering
\resizebox{0.50\textwidth}{!}{%
\begin{tabular}{cccc}
\toprule
 {\bf Model} & {\bf emrQA F1} & {\bf Clift F1} & \textbf{Gap ($\delta$})\\
\midrule
   BERT-ClinicalQA & 88.28 &          17.31 & 70.97 \\
         BioBert-B & 87.55 &          17.38 & 70.17 \\
   BioClinicalBert & 86.98 &          17.67 & 69.32 \\
    BlueBert-MIMIC & 89.16 &          19.20 & 69.96 \\
   Electra-Squad-L & 86.98 &          19.44 & 67.55 \\
   BlueBert-PubMed & 86.98 &          19.81 & 67.18 \\
DistilBert-squad-B & 85.68 &          20.45 & 65.23 \\
        PubMedBERT & 91.72 &          21.95 & 69.77 \\
   BioBert-squad-L & 90.95 &          22.53 & 68.42 \\
        Bert-Large & 88.81 &          23.25 & 65.56 \\
    Biomed-Roberta & 89.76 &          23.59 & 66.18 \\
   Roberta-squad-B & 89.84 &          24.99 & 64.85 \\
\bottomrule
\end{tabular}}
\caption{Performance of different models on the emrQA Testset and new Medication Testset}
\label{tab:medicationtab}
\end{table*}

\begin{table*}
\small
\centering
\resizebox{0.50\textwidth}{!}{%
\begin{tabular}{cccc}
\toprule
 {\bf Model} & {\bf emrQA F1} & {\bf Clift F1} & \textbf{Gap ($\delta$})\\
\midrule
PubMedBERT & 91.72 &     21.35 & 70.37 \\
    Biomed-Roberta & 89.76 &     22.72 & 67.04 \\
         BioBert-B & 87.55 &     23.37 & 64.18 \\
   Electra-Squad-L & 86.98 &     24.09 & 62.89 \\
   BioClinicalBert & 86.98 &     24.51 & 62.47 \\
    BlueBert-MIMIC & 89.16 &     25.50 & 63.67 \\
   BioBert-squad-L & 90.95 &     25.87 & 65.08 \\
        Bert-Large & 88.81 &     26.17 & 62.63 \\
   BERT-ClinicalQA & 88.28 &     27.04 & 61.24 \\
DistilBert-squad-B & 85.68 &     27.10 & 58.58 \\
   BlueBert-PubMed & 86.98 &     27.54 & 59.44 \\
   Roberta-squad-B & 89.84 &     29.90 & 59.94 \\
\bottomrule
\end{tabular}}
\caption{Performance of different models on the emrQA Testset and new Heart Testset}
\label{tab:hearttab}
\end{table*}

\begin{table*}
\small
\centering
\resizebox{0.50\textwidth}{!}{%
\begin{tabular}{cccc}
\toprule
 {\bf Model} & {\bf emrQA F1} & {\bf Clift F1} & \textbf{Gap ($\delta$})\\
\midrule
Biomed-Roberta & 89.76 &      25.54 & 64.22 \\
        PubMedBERT & 91.72 &      26.86 & 64.86 \\
DistilBert-squad-B & 85.68 &      29.19 & 56.49 \\
         BioBert-B & 87.55 &      30.31 & 57.24 \\
   BERT-ClinicalQA & 88.28 &      32.45 & 55.83 \\
   BioClinicalBert & 86.98 &      33.46 & 53.52 \\
   Roberta-squad-B & 89.84 &      33.73 & 56.11 \\
    BlueBert-MIMIC & 89.16 &      36.49 & 52.67 \\
   BlueBert-PubMed & 86.98 &      36.51 & 50.47 \\
   BioBert-squad-L & 90.95 &      36.56 & 54.40 \\
   Electra-Squad-L & 86.98 &      36.90 & 50.08 \\
        Bert-Large & 88.81 &      37.31 & 51.49 \\
\bottomrule
\end{tabular}}
\caption{Performance of different models on the emrQA Testset and new Cancer Testset}
\label{tab:cancertab}
\end{table*}

\begin{table*}
\small
\centering
\resizebox{0.50\textwidth}{!}{%
\begin{tabular}{cccc}
\toprule
 {\bf Model} & {\bf emrQA F1} & {\bf Clift F1} & \textbf{Gap ($\delta$})\\
\midrule
DistilBert-squad-B & 85.68 &       12.62 & 73.07 \\
         BioBert-B & 87.55 &       12.89 & 74.66 \\
   BERT-ClinicalQA & 88.28 &       13.00 & 75.28 \\
    BlueBert-MIMIC & 89.16 &       13.60 & 75.56 \\
   BlueBert-PubMed & 86.98 &       14.07 & 72.91 \\
   BioClinicalBert & 86.98 &       14.20 & 72.78 \\
   BioBert-squad-L & 90.95 &       14.49 & 76.46 \\
        PubMedBERT & 91.72 &       14.65 & 77.08 \\
   Electra-Squad-L & 86.98 &       14.81 & 72.17 \\
        Bert-Large & 88.81 &       16.92 & 71.88 \\
    Biomed-Roberta & 89.76 &       17.31 & 72.46 \\
   Roberta-squad-B & 89.84 &       19.88 & 69.96 \\
\bottomrule
\end{tabular}}
\caption{Performance of different models on the emrQA Testset and new Obesity Testset}
\label{tab:obesitytab}
\end{table*}

\begin{table*}
\small
\centering
\resizebox{0.50\textwidth}{!}{%
\begin{tabular}{cccc}
\toprule
 {\bf Model} & {\bf emrQA F1} & {\bf Clift F1} & \textbf{Gap ($\delta$})\\
\midrule
BioBert-B & 87.55 &     32.57 & 54.98 \\
   BERT-ClinicalQA & 88.28 &     34.31 & 53.97 \\
   Electra-Squad-L & 86.98 &     35.13 & 51.85 \\
DistilBert-squad-B & 85.68 &     36.18 & 49.51 \\
   BioClinicalBert & 86.98 &     36.90 & 50.08 \\
    Biomed-Roberta & 89.76 &     38.38 & 51.39 \\
   BlueBert-PubMed & 86.98 &     40.11 & 46.87 \\
        Bert-Large & 88.81 &     42.84 & 45.97 \\
   BioBert-squad-L & 90.95 &     43.36 & 47.60 \\
        PubMedBERT & 91.72 &     44.75 & 46.97 \\
   Roberta-squad-B & 89.84 &     45.16 & 44.68 \\
    BlueBert-MIMIC & 89.16 &     45.34 & 43.82 \\
\bottomrule
\end{tabular}}
\caption{Performance of different models on the emrQA Testset and new Smoke Testset}
\label{tab:smoketab}
\end{table*}

\begin{figure}[htp]

\includegraphics[width=.5\textwidth]{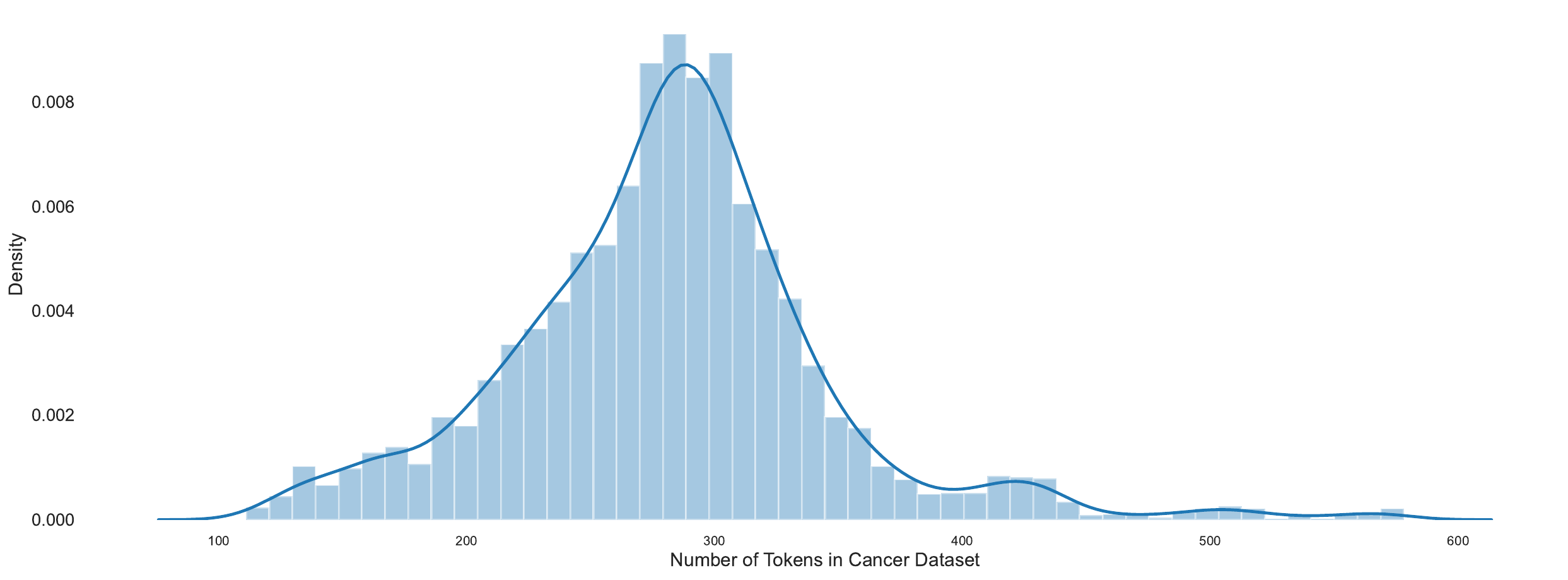}\quad
\includegraphics[width=.5\textwidth]{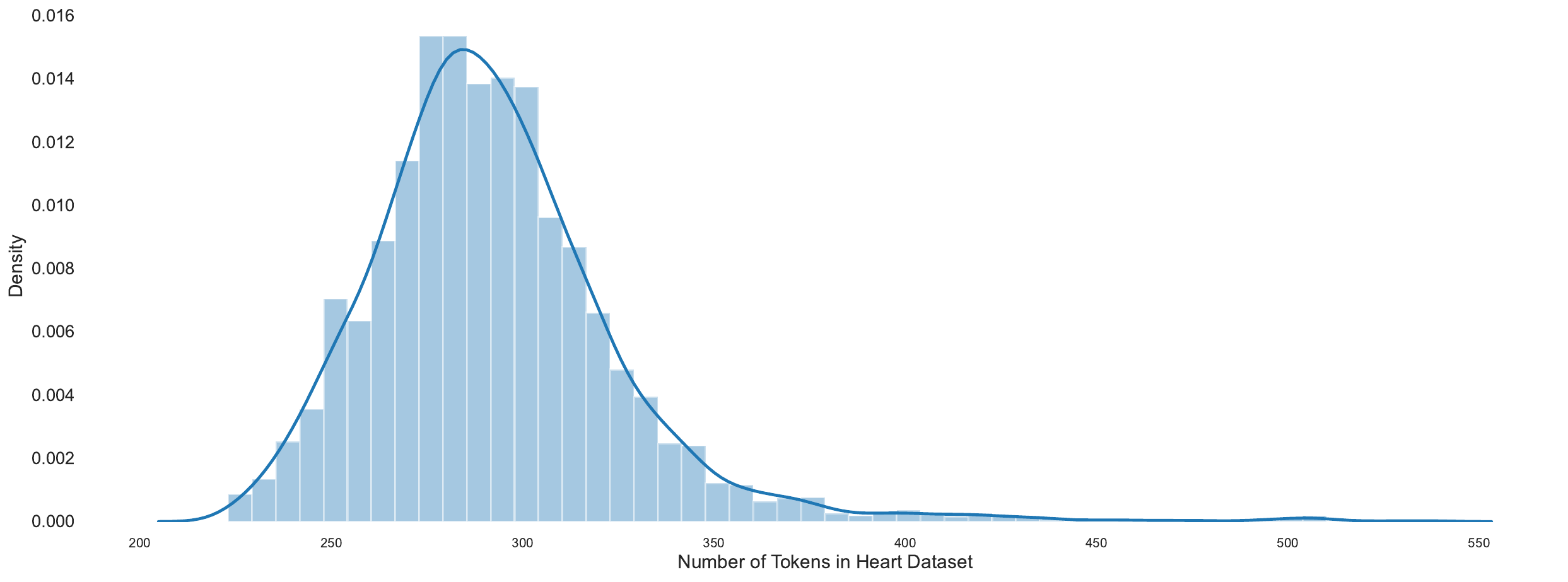}\quad
\includegraphics[width=.5\textwidth]{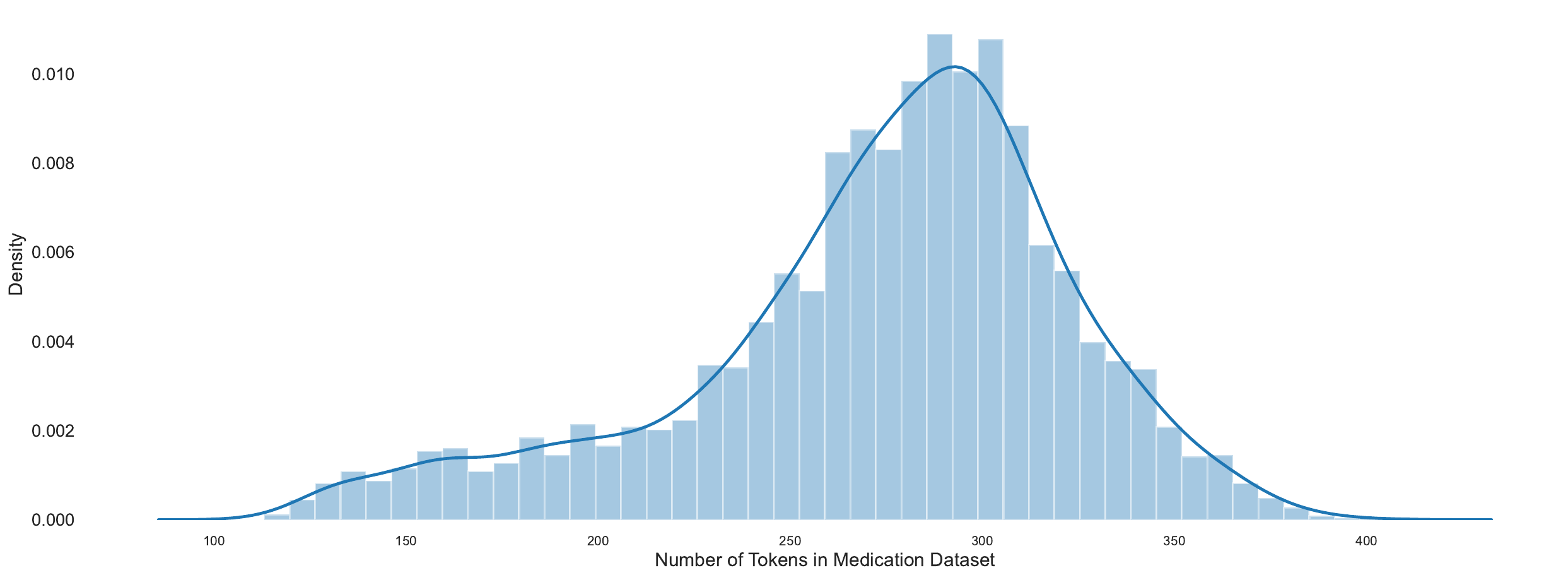}

\medskip

\includegraphics[width=.5\textwidth]{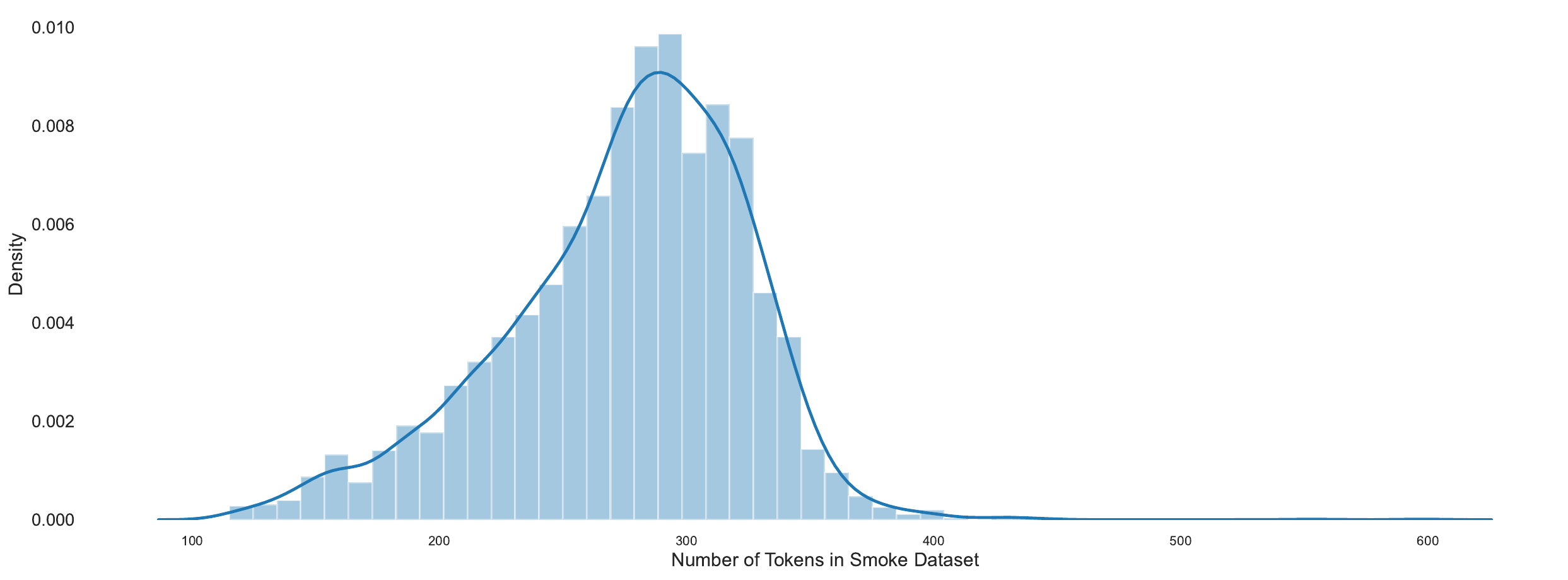}\quad
\includegraphics[width=.5\textwidth]{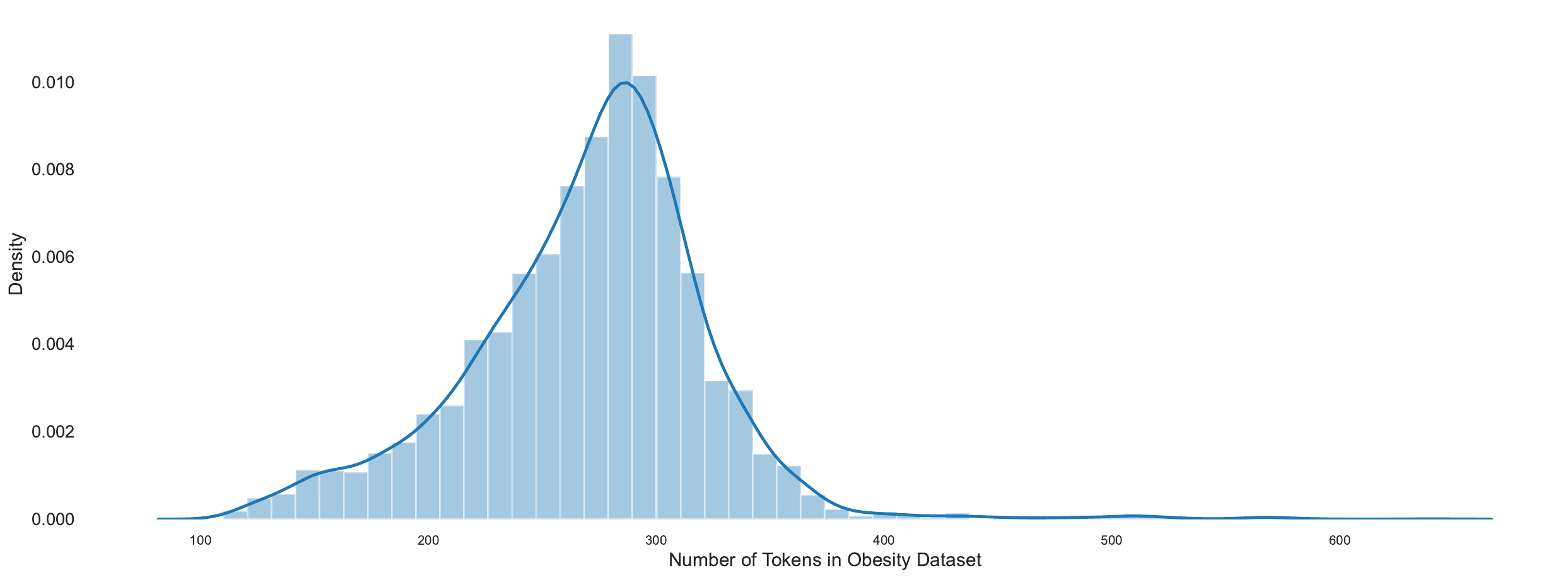}

\caption{Subfigure (a), (b), (c), (d), and (e) shows the Distribution of the context length in the Cancer, Heart, Medication, Smoke and Obesity dataset, respectively}
\label{fig:tokendist}
\end{figure}

\end{document}